\documentclass[a4paper,twoside]{article}
\usepackage[table,xcdraw]{xcolor}
\usepackage{epsfig}
\usepackage{subcaption}
\usepackage{calc}
\usepackage{amssymb}
\usepackage{amstext}
\usepackage{url}
\usepackage{graphicx}
\usepackage{amsmath}
\usepackage{amsthm}
\usepackage{multicol}
\usepackage{pslatex}
\usepackage{apalike}
\usepackage{SCITEPRESS} 
\newtheorem{theorem}{Theorem}[section]

\newtheorem{proposition}{Proposition}[section]
\newtheorem{lemma}[theorem]{Lemma}
\newtheorem{definition}{Definition}[section]

\begin{document}

\title{Reinforcement Learning Guided by Provable Normative Compliance}

\author{\authorname{Emery A. Neufeld\orcidAuthor{0000-0001-5998-3273}}
\affiliation{Faculty of Informatics, TU Wien, Austria}
\email{emeric.neufeld@tuwien.ac.at}
}


\abstract{Reinforcement learning (RL) has shown promise as a tool for engineering safe, ethical, or legal behaviour in autonomous agents. Its use typically relies on assigning punishments to state-action pairs that constitute unsafe or unethical choices. Despite this assignment being a crucial step in this approach, however, there has been limited discussion on generalizing the process of selecting punishments and deciding where to apply them. In this paper, we 
adopt an approach that leverages an existing framework -- the normative supervisor of \cite{NBCG2021} -- during training. This normative supervisor is used to dynamically translate states and the applicable normative system into defeasible deontic logic theories, feed these theories to a theorem prover, and use the conclusions derived to decide whether or not to assign a punishment to the agent. We use multi-objective RL (MORL) to balance the ethical objective 
of avoiding violations with a non-ethical objective; we will demonstrate that our approach works for a multiplicity of MORL techniques, and show that it is effective regardless of the magnitude of the punishment we assign.}

\onecolumn \maketitle \normalsize \setcounter{footnote}{0} \vfill
\section{Introduction}
\label{sec:introduction}
An increasing amount of attention is being devoted to the design and implementation of ethical and safe autonomous agents. Among this growing body of research are multiple approaches adapting reinforcement learning (RL) methods to learn compliance with ethical standards. \cite{RLR2021} notes that 
these approaches can be split into two main tasks: reward specification (the encoding of ethical information into a reward function) and ethical embedding (incorporating rewards into the agent's learning environment). In \cite{RLR2021}, the authors focus on the latter; here, we will more closely work with the former. We can identify two essential challenges with reward specification: we must decide (1) when to assign rewards or punishments to the agent, and (2) what their magnitude should be. In this paper, we address these challenges through a mechanism for identifying, and assigning punishments for, behaviours that do not comply with a given ethical or legal framework presented as a normative system. 

Normative systems, as we discuss them here, consist of sets of norms, some of which govern the behaviour of an agent (regulative norms), while others define institutional facts within the system (constitutive or counts-as norms). These norms can be combined in complex ways. 
A familiar example can be found in driving regulations. Though we might have a simple rule like ``do not travel above 50 kph", we might also have more complex regulations along the lines of: ``(1) you are forbidden from proceeding at a red light; however, (2) you are permitted to turn right at a red light, if it is safe to do so". In this example, we have a simple prohibition -- ``you are forbidden from proceeding at a red light" -- but then we have an exception to that rule in the form of a conditional permission, which is to be applied ``if it is safe to do so". Though a human being might be able to intuit that this means that they can turn right at a red light only if it will not cause a collision of any kind, an automated vehicle would not have these intuitions. 
To recognize that (2) is an exception to (1), we might have a constitutive norm designating that ``turning right counts as proceeding". We will furthermore want to address the concept of safety, and assert perhaps ``moving within 1 m of another object counts as unsafe", defining one -- of possibly many -- conditions for safety. Thus, in this more complex case, we have a prohibition, a conditional permission, and two constitutive norms, all of which work together to define a given action in a given state as compliant, or not. 

The typical approach to teaching RL agents safe or ethical behaviour is to assign punishments to non-compliant state-action pairs, but this will not always be a simple matter, in the context of a large or complex normative system. 
Our approach is to model the normative system with a formal language tailored to representing and deriving conclusions from such systems, and use a theorem prover to check if a state-action pair is compliant with the system. If it is not, a punishment is incurred by the agent. The medium through which we automate this process is a \textit{normative supervisor}.


In \cite{NBCG2021}, a normative supervisor for RL agents employing a defeasible deontic logic theorem prover is introduced to address the challenge of facilitating transparent normative reasoning in learning agents. 
This external module works by filtering out non-compliant actions, allowing the agent to choose the best action out of all compliant actions; if there are no compliant actions, the supervisor possesses a ``lesser evil" submodule that can score actions based on how much of the normative system they violate, allowing the supervisor to recommend only minimally non-compliant actions to the agent. In this paper, we describe how the supervisor can be used to evaluate the compliance of a given action, allowing the supervisor to guide the agent more indirectly, by simply evaluating the agent's actions instead of restricting them, providing feedback rather than instructions. This feedback will come in the form of punishments for a reinforcement learning agent, which we will show are effective in molding the agent's behaviour regardless of their magnitude. 
We adopt multi-objective reinforcement learning (MORL) as a framework for learning policies that accomplish both the agent's original goal and its compliance to a normative system, as discussed in in \cite{VDFFM2018} and \cite{RLR2021}.

We demonstrate the efficacy of this approach on an RL agent learning to play a variation of the game Pac-Man. We employ the ``Vegan Pac-Man" case study utilized in \cite{N2019} and \cite{NBCG2021}, adapting it into a more complex normative system that characterizes a concept of ``benevolence" for Pac-Man and mandates adherence to it. 
Our results show that our more complex formulation manifests in the same behaviour as the ``vegan" Pac-Man from the above papers -- a version of the game where Pac-Man is forbidden from eating ghosts, even though doing so would increase its score in the game, thus creating two competing objectives, one ethical, and one unethical. 
The key difference here lies in how the mandate for this behaviour is formulated. 
 
We engineer this desired behaviour with two approaches to single-policy MORL: the linear scalarization MORL approach used in \cite{RLR2021}, and the ranked MORL first presented in \cite{GKS1998} and later simplified in \cite{VDBID2011} as thresholded lexicographic Q-learning (TLQ-learning). Our results show these two techniques producing the same optimal policy. After demonstrating these techniques on a simplified version of the game Pac-Man, we turn our attention to their potential adaptations of our techniques to more complex environments and normative systems and the potential challenges that arise therefrom.

\subsection{Related Work}
There is a wealth of research on \textit{safe} RL. A broad survey of safe reinforcement learning techniques can be found in \cite{GF2015}, but since this paper was published, significant research has been done on the topic. 
Among this research are several logic-based frameworks. Using linear temporal logic (LTL) specifications to constrain an agent's actions is one of the most common approaches. For instance, in \cite{ABEKNT2018} and \cite{JKJSB2020}, a shield is synthesized from LTL specifications and abstractions of the environment, which prevents the agent from moving into unsafe states. In \cite{HAK2018}, \cite{HKAKPL2019}, and  \cite{HAK2020} the authors translate LTL specifications into automata and learn a policy based on a reward function defined by the non-accepting states of the automata. In \cite{KS2018}, LTL specifications are similarly translated into automata which are tailored for direct conflict resolution between specifications. \cite{KS2018} is the only one of these approaches that explicitly addresses normative reasoning, and it restricts itself to solving conflicts between strictly weighted prescriptive norms. Other subtleties of normative reasoning 
are not so simple to represent with LTL; even if we can manually transform a normative system into a set of LTL constraints, doing so introduces the possibility of human error while translating complex dynamics of normative reasoning into LTL, which may not be able to express these dynamics in a natural way. 
LTL is ideal for straightforward safety constraints, but less well suited for legal or ethical norms \cite{GH2015}.

Nevertheless, RL is not an uncommon way to address the learning of ethical behaviour in autonomous agents. Many examples of this have popped up in recent years, such as the framework outlined in \cite{AML2016}, the reward-shaping approach in \cite{WL2018}, or the MORL with contextual multi-armed bandits presented in \cite{N2019} and \cite{BBMR2019}. More recently, 
\cite{RLR2021} used an ``ethical multi-objective Markov decision process" to design an ethical environment for teaching an agent ethical behaviour. While all the above approaches to ethically-compliant agents include assigning rewards and punishments to the actions taken by an agent in order to induce compliant behaviour, there is limited discussion as to how or why the rewards are assigned where they are. Additionally, in work focusing on reinforcement learning, there has been a lack of consideration that the agent may not be subject to simple constraints on behaviour; an agent might be subject to an entire normative system, where counts-as relations or constraints exist, as discussed in \cite{BT2003}. 
It may not be immediately obvious what situations are compliant with a normative system and which are not; often we require contextual definitions to make sense of normative systems, and we may not be able to give an exhaustive description of all instances that qualify as non-compliant.  Though there are undoubtedly limits to using reinforcement learning to model compliant behaviour in such systems, it remains to be seen how well behaviour resulting from such kinds of reasoning can be learned with RL.

\section{Background}
In this section we review some background topics that will be necessary building blocks of our approach to learning compliant behaviour. 
First, we summarize some key definitions in RL and MORL, and then discuss normative systems. We then further focus our discussion by defining a formal language for normative reasoning (defeasible deontic logic \cite{GORS2013,jaamas:bio}), and the normative supervisor \cite{NBCG2021} that utilizes its theorem prover SPINdle \cite{ruleml09:spindle}.

\subsection{Multi-Objective Reinforcement Learning}
The underlying environment of a reinforcement learning problem is formalized as a Markov decision process (MDP), defined below:
\begin{definition}
An MDP is a tuple $$\langle S, A, R, P\rangle$$
where $S$ is a set of states, $A$ is a set of actions, $R:S\times A\to \mathbb{R}$ is a scalar reward function over states and actions, and $P:S\times A\times S\to [0,1]$ is a probability function that gives the probability $P(s,a,s')$ of transitioning from state $s$ to state $s'$ after performing action $a$.
\end{definition}

The goal of reinforcement learning is to find a policy $\pi:S\to A$ which designates optimal behaviour;
this optimality is determined with respect to a value function defined as:
$$
    V^{\pi}(s)=E\left[\sum^{\infty}_{t=0}\gamma^{t} r_{i+t+1} | s_i=s\right]
$$
which represents the expected accumulated value onward from state $s$ if policy $\pi$ is followed. In the above function, $r_{t}$ is the reward earned from the reward function $R$ at timestep $t$ and $\gamma\in [0,1]$ is a discount factor (so that rewards in the future do not have as much weight as the current reward). We can similarly define a Q-function: 
$$
    Q^{\pi}(s,a)=E\left[\sum^{\infty}_{t=0}\gamma^{t} r_{i+t+1} | s_i=s, a_i=a\right]
$$
which predicts the expected cumulative reward from $R$ given that the agent is in state $s$ taking action $a$. 

The goal of RL, then, is to find an optimal policy $\pi^{*}$ such that $$V^{\pi^{*}}(s) = \max_{\pi\in \Pi}V^{\pi}(s)$$
where $\Pi$ is the set of all policies over the MDP. This is accomplished by learning a Q-function such that $\pi^{*}(s)\in\text{argmax}_{a\in A}Q(s,a)$.

Multi-objective RL (MORL) differs from regular RL only in that instead of learning over an MDP, we want to learn over a multi-objective MDP (MOMDP); an MDP that has instead of a single reward function $R$, multiple reward functions $\vec{R}=(R_{1},...,R_{n})^T$, each corresponding to a different objective, and therefore a different value function. 

In other words, MOMDPs differ from MDPs only in that instead of a single scalar reward function, they utilize a vector of reward functions. In turn, we can define a vector of Q-functions $\vec{Q}=(Q_{1},...Q_{n})^T$, where for each $1\le i\le n$, $Q_{i}:S\times A\to \mathbb{R}$ predicts the expected cumulative reward from $R_{i}$ given that the agent is in state $s$ taking action $a$. During training, each one of these Q-functions is updated with its corresponding reward function.

Unfortunately, the existence of multiple objectives results in a more complex set of semi-optimal policies; one policy might maximize rewards from $R_{i}$ but not $R_{j}$, or the opposite could be the case. This is especially a problem with competing objectives. This is where we get the notion of \textit{Pareto dominance}. A policy strongly dominates others if it results in superior outcomes for all objectives. However, in some cases, like the case where there are competing objectives, no such policy may exist. In this case we look at whether one policy \textit{weakly} dominates another second policy; that is, the first policy results in superior outcomes with respect to one objective, and is not inferior with respective to any of the other objectives to the second policy. If two policies weakly dominate each other over at least one objective, they are called incomparable.

What we are interested in is the \textit{Pareto front}: if we remove all strictly dominated policies, the only policies left form the Pareto front, which is the set of all dominant or incomparable policies. The task of MORL is to find policies in this Pareto front. There are a plethora of methods for choosing an action in the presence of competing objectives, two of which will be discussed in Sect.~3.1 and 3.2.


\subsection{Normative Reasoning}
In this paper our goal is to influence the learning of an agent with normative reasoning. 
Normative reasoning differs from classical logical reasoning in that its focus is not only on the truth or falsity of a given statement, but also the application of the modality of obligation -- and the related modalities of permission and prohibition -- to it. Normative reasoning introduces challenging dynamics that cannot be effectively captured by classical logic. There is debate as to the nature of these dynamics and what tools can or should be used to model them \cite{BCEGG2013}; suggested crucial characteristics of normative reasoning include the capability to represent both constitutive and regulative norms and defeasibility \cite{PGRTBP2011}.

As noted above, normative reasoning often entails dealing with two types of norms: constitutive and regulative norms (see \cite{BT2004}). 
Regulative norms describe the obligations, prohibitions and permissions an agent is subject to. Typically, these are represented as: $\textbf{O}(p|q)$ ($p$ is obligatory when $q$), $\textbf{F}(p|q)$ ($p$ is forbidden when $q$), and $\textbf{P}(p|q)$ ($p$ is strongly permissible when $q$). Prohibition can be seen as merely the obligation of a negative, i.e., $\textbf{F}(p|q)\equiv \textbf{O}(\neg p|q)$, and strong permission occurs when we express the permissiveness of an action explicitly, and this explicit permission overrides the prohibition of such an action. In other words, strong permissions work as exceptions to prohibitions and obligations -- this is one place where the defeasibility condition mentioned in the above paragraph comes in.

On the other hand, constitutive norms typically take the form “in context c, 
concept A counts as concept B" \cite{JS1996}, where A generally refers to a more concrete concept (e.g., turning right) and B to a more abstract one (e.g., proceeding). When we have a constitutive norm that informs us that $x$ counts as $y$, we will here denote it as $\textbf{C}(x, y)$. Constitutive norms can be used to define more complex concepts like ``safety" or ``benevolence" within a specific normative system:
\begin{definition}
A normative system is a tuple $\mathcal{N}=\langle \mathcal{C}, \mathcal{R}, >\rangle$ where $\mathcal{C}$ is a set of constitutive norms, $\mathcal{R}$ is a set of regulative norms, and $>$ is a priority relation that resolves conflicts between norms, should they exist.
\end{definition}
Below, we describe a formal language that can model the dynamics of such normative systems.

\subsubsection{Defeasible Deontic Logic}
Defeasible deontic logic (DDL) is a computationally feasible, albeit expressive logic that addresses the challenges with normative reasoning discussed above.

DDL facilitates reasoning over literals (propositional atoms $p$ and their negations $\neg p$), modal literals (literals subject to a modality, for example obligation $\textbf{O}(p)$), and rules defined over them. Rules can be strict ($\to$), defeasible ($\Rightarrow$), or defeaters ($\rightsquigarrow$). In strict rules, the consequent strictly follows from the antecedent without exception, whereas the consequents of defeasible rules \textit{typically} follow from the antecedent, unless there is evidence to the contrary. This evidence can come in the form of conflicting rules or defeaters, which cannot be used to derive a conclusion; rather, they prevent a conclusion from being reached by a defeasible rule. Rules can be constitutive (bearing the subscript $C$) or regulative (bearing the subscript $O$)

To provide the examples of the above types of rules, consider the constitutive norm $\textbf{C}(x,y)$ which holds without fail. This will be expressed as $$x\to_{C}y$$ Also consider the prohibition $\textbf{F}(p|q_1)$; this can be expressed as $$q_{1}\Rightarrow_{O}\neg p$$
A strong permission acting as an exception to the above prohibition, $\textbf{P}(p|q_{2})$, can be expressed as $$q_{2}\rightsquigarrow_{O} p$$

If we have a collection of facts $F$ and a normative system (sets of constitutive and regulative norms) defined in the language above, we can form a defeasible theory:
\begin{definition}[Defeasible Theory \cite{G2018}]
A defeasible theory $D$ is defined by the tuple $\langle F, R_{O}, R_{C}, >\rangle$, where $F$ is a set of facts in the form of literals, $R_{O}$ is a set of regulative rules, $R_{C}$ is a set of constitutive rules, and $>$ is a superiority relation over conflicting rules.
\end{definition}

The theorem prover for DDL, SPINdle \cite{ruleml09:spindle} takes a defeasible theory and outputs a set of conclusions, which is the set of literals occurring in the defeasible theory tagged according to their status as provable or not provable. There are several types of conclusions we can derive from a defeasible theory; conclusions can be \textit{negative} or \textit{positive},  \textit{definite} or \textit{defeasible}, or \textit{factual} or \textit{deontic}. 
The proof tags are: \begin{itemize}
    \item $+\Delta_{*}$:  definitely provable conclusions are either facts or derived only from strict rules and facts.
    \item $-\Delta_{*}$: definitely refutable conclusions are neither facts nor derived from only strict rules and facts.
    \item $+\partial_{*}$: defeasibly provable conclusions are not refuted by any facts or conflicting rules, and are implied by some undefeated rule.
    \item $-\partial_{*}$: defeasibly refutable conclusions are conclusions for which their complemented literal is defeasibly provable, or an exhaustive search for a constructive proof for the literal fails. 
\end{itemize}
\cite{G2018}

Note that for factual conclusions, $* := C$, and for deontic conclusions, $* := O$. So, for example, if we were to derive ``$p$ is forbidden" from a defeasible theory $D$, we would have $D\vdash +\Delta_{O} \neg p$ or $D\vdash +\partial_{O}\neg p$, depending on whether or not the conclusion was reached definitely or defeasibly. These conclusions can be computed in linear time \cite{GORS2013}.

\subsubsection{The Normative Supervisor}
\label{sect:super}
The normative supervisor -- introduced in \cite{NBCG2021} -- is an external reasoning module capable of interfacing with a reinforcement learning agent. It is composed primarily of a normative system, an automatic translation submodule to transform facts about the environment and the normative system into a defeasible theory, and a theorem prover to reason about said defeasible theory. The logic and associated theorem prover used both in \cite{NBCG2021} and here are DDL and SPINdle respectively.

The normative supervisor monitors the agent and its environment, dynamically translating states and the normative system into defeasible theories; generally, facts about the environment (e.g., the locations of objects the agent observes) are translated to literals, and compose the set of facts of the defeasible theory. Note that to perform this translation, we do not need to have any kind of model of the environment (e.g., we do not need to have learned an MDP), we only need, for example, a simple labelling function that maps states to atomic propositions which are true in that state.

Meanwhile, the constitutive norms and regulative norms are translated into DDL rules and added to the defeasible theory. The final addition to the defeasible theory will be sets of \textit{non-concurrence constraints}:
$$\textbf{C}(a, \neg a')\in \mathcal{C} \text{, }\forall a'\in A-\{a\}$$
These constraints establish that only one action can be taken; if we have an obligation of $a$, it effectively forbids the agent from choosing another action.

We will use $Th(s, \mathcal{N})$ to denote the theory generated for state $s$, the normative system $\mathcal{N}$, and the set of non-concurrence constraints. Based on the output of the theorem prover, the supervisor filters out actions that do not comply with the applicable normative system, and constructs a \textit{set of compliant solutions} (see \cite{NBCG2021} for a full description). 

However, we are not here interested in finding an entire set of compliant solutions. Our task is simpler; given an action $a$ and a current state $s$, we only want to know if the prohibition of $a$ is provable from the theory; that is, if $+\partial_{O}\neg a$ is output by SPINdle given the input $Th(s,\mathcal{N})$. 
This procedure is much simpler than the supervisor's main function, as we only need to search a set of conclusions for a single specific conclusion, instead of constructing an entire set of compliant solutions from all conclusions given.

\section{Norm-Guided Reinforcement Learning}
\label{sect:ngql}
Here we discuss how the mechanisms of the normative supervisor explained in Sect.~\ref{sect:super} can be utilized within a MORL framework to learn compliant behaviour, in a process we will call norm-guided reinforcement learning (NGRL). NGRL is a customizable approach to implementing ethically compliant behaviour, that blends techniques and tools from both logic and reinforcement learning.

The basic approach is this: given an agent with an objective $x$ (and an associated reward function $R_x(s,a)$), we define a second reward function that assigns punishments when an agent violates a normative system $\mathcal{N}$. We call this second reward function a \textit{non-compliance function}:

\begin{definition}[Non-Compliance Function]
A non-compliance function for the normative system $\mathcal{N}$ is any function of the form:
$$
R_{\mathcal{N},p}(s, a) = \begin{cases} 
p & Th(s, \mathcal{N})\vdash +\partial_{O}\neg a \\
      0 & \text{otherwise}
   \end{cases}
$$
where $p\in \mathbb{R}^{-}$. 
\end{definition}
In the above definition, $p$ is called the \textit{penalty}, and $Th(s, \mathcal{N})$ is a normative system translated together with a state $s$ into a defeasible theory, for example by the translators used by the normative supervisor.  The automated derivation of conclusions from $Th(s, \mathcal{N})$ solves the first challenge of reward specification, allowing us to dynamically determine the compliance of an action in a given state and thereby assign a punishment. As for the second challenge, we will discuss the magnitude of $p$ in Sect.~3.3. 
Summarily, what this non-compliance function $R_{\mathcal{N},p}$ does is, for each state-action pair $(s,a)$, assign a fixed punishment if and only if $a$ violates $\mathcal{N}$ in state $s$. 


Now, suppose we have an agent we want to learn objective $x$ with the reward function $R_{x}(s,a)$; it can learn to do so over the MDP 
$$\mathcal{M} = \langle S, A, R_{x}, P\rangle$$ 
If we have a normative system $\mathcal{N}$ we wish to have the agent adhere to while it fulfills objective $x$, we can build an MOMDP we will call a \textit{compliance MDP}:
\begin{definition}[Compliance MDP]
Suppose we have a single-objective MDP $\mathcal{M} = \langle S, A, R_{x}, P\rangle$. Then we can create an associated compliance MDP by introducing a non-compliance function $R_{\mathcal{N},p}(s, a)$ to form the MOMDP:
$$\mathcal{M}_{\mathcal{N},p}=\langle S, A, (R_{x}, R_{\mathcal{N},p})^{T}, P\rangle$$
\end{definition}

The goal of NGRL is to find an optimal-ethical policy for a compliance MDP, where the terminology \textit{optimal-ethical} is taken from  \cite{RLR2021}. We adapt their definitions below:
\begin{definition}[Ethical Policy \cite{RLR2021}]
Let $\Pi$ be the set of all policies over the compliance MDP $\mathcal{M}_{\mathcal{N},p}$. A policy $\pi^{*}\in \Pi$ is ethical if and only if it is optimal with respect to the value function $V^{\pi}_{\mathcal{N},p}$ corresponding to $R_{\mathcal{N},p}$. In other words, $\pi^{*}$ is ethical for $\mathcal{M}_{\mathcal{N},p}$ iff
\begin{equation}
    V^{\pi^{*}}_{\mathcal{N},p}(s)=\max_{\pi\in \Pi}V^{\pi}_{\mathcal{N},p}(s)
\end{equation}
for all $s$.
\end{definition}

\begin{definition}[Ethical-Optimal Policy \cite{RLR2021}]
Let $\Pi_{\mathcal{N}}$ be the set of all ethical policies for $\mathcal{M}_{\mathcal{N},p}$. Then $\pi^{*}\in \Pi_{\mathcal{N}}$ is ethical-optimal for $\mathcal{M}_{\mathcal{N},p}$ iff
\begin{equation}
    V_{x}^{\pi^{*}}(s)=\max_{\pi\in \Pi_{\mathcal{N}}}V^{\pi}_{x}(s)
\end{equation}
for all $s$.
\end{definition}

There are a plethora of MORL techniques for learning optimal solutions for MOMDPs; there will be multiple ways to learn ethical-optimal solutions for some compliance MDP $\mathcal{M}_{\mathcal{N},p}$. Below we will discuss two different MORL algorithms that we adopt to solve $\mathcal{M}_{\mathcal{N},p}$ for ethical-optimal policies. After defining these methods we will prove some useful properties of NGRL and discuss some notable limitations.

\subsection{Linear Scalarization}
A common approach to MORL is scalarizing the vector of Q-functions associated with the multiplicity of objectives by weighting each function with positive values \cite{RVWD2013}. Here, we will use an approach similar to \cite{RLR2021}; however, we simplify their case slightly by having only one function associated with ethical behaviour, $R_{\mathcal{N},p}$. 

Given a weight vector $\vec{w}\in \mathbb{R}_{n}^{+}$, the approach we take is to learn $Q_{\mathcal{N},p}(s,a)$ and $Q_x(s,a)$ concurrently, and scalarize the vector of Q-functions via the inner product of the weight and Q-value vector. In other words, we want to maximize
\begin{equation}
    V_{scalar}(s)=\vec{w}\cdot \vec{V}(s)
\end{equation}
where $\vec{V}(s)=(V_x(s), V_{\mathcal{N},p}(s))^{T}$, by selecting actions through the scalarized Q-function be defined as
\begin{equation}
    Q_{scalar}(s,a)=\vec{w}\cdot \vec{Q}(s,a)
\end{equation}
where $\vec{Q}(s,a)=(Q_x(s,a), Q_{\mathcal{N},p}(s,a))^{T}$. In the case of the compliance MDP $\mathcal{M}_{\mathcal{N},p}$ defined above, we can assume without loss of generality that this weight vector is of the form $(1, w)^{T}$ for some $w\in \mathbb{R}^{+}$ \cite{RLR2021}.

How do we find an appropriate $w$? \cite{RLR2021} describes a method using Convex Hull Value Iteration \cite{BN2008}. For the sake of brevity, we will not discuss this algorithm in detail, and only relay the results from \cite{RLR2021}:

\begin{proposition}
If at least one ethical policy exists, there exists a value $w\in \mathbb{R}^{+}$ such that any policy maximizing $V_{scalar}(s)$ is ethical-optimal. 
\end{proposition}
\noindent \textit{Proof sketch:} This proposition follows from Theorem 1 of \cite{RLR2021}. \qedsymbol

\subsection{TLQ-Learning}
 \cite{GKS1998} describes a MORL approach where certain objectives can be prioritized over others. \cite{VDBID2011} gives what they describe as a naive approach to the work in \cite{GKS1998}, which we will follow here.

This approach is tailored in particular to problems where there is a single objective  that must be maximized, while all other objectives don't need to be maximized, as such, but rather must satisfy a given threshold. For us, implementing this framework is quite simple, as we have only two objectives: the objective represented by $R_x$ (which we want to maximize) and the objective represented by $R_{\mathcal{N},p}$, which we would like to prioritize and constrain. Thus, we need to set a constant threshold $C_{\mathcal{N},p}$ for the objective represented by $R_{\mathcal{N},p}$, while the threshold for the objective represented by $R_{x}$ is set to $C_x=+\infty$.

When we select an action, instead of Q-values, we consider CQ-values defined as
$$CQ_{i}(s,a)=\min (Q_{i}(s,a), C_{i})$$
which takes the thresholds into account. The next step in this approach is to order actions based on their CQ-values. Again, since we only have two objectives, we were able to simplify the procedure given in \cite{VDBID2011} to:\begin{enumerate}
    \item Create a set $eth(s) =\{a\in A|CQ_{\mathcal{N},p}(s,a) = max_{a'\in A}CQ_{\mathcal{N},p}(s,a')\}$ of actions with the highest $CQ_{\mathcal{N},p}$-values for the current state $s$.
    \item Return a subset $opt(s)=\{a\in eth(s) | CQ_x(s,a) = max_{a'\in eth(s)}CQ_x(s,a')\}$ of the actions of $eth(s)$, with the highest $CQ_{x}$-values. 
\end{enumerate}
Our policy must select, for a state $s$, an action from $opt(s)$.

 Summarily, we always take an action that is maximal for $CQ_{\mathcal{N},p}$, and within the actions that maximize $CQ_{\mathcal{N},p}$, we take the action with a value that is maximal for $Q_{x}$.

\begin{proposition}
If an ethical policy exists, when $C_{\mathcal{N},p}=0$, any policy such that $\pi(s)\in opt(s)$ for all $s$ designates an ethical-optimal policy for $\mathcal{M}_{\mathcal{N},p}$.
\end{proposition}
\noindent \textit{Proof sketch:} Since the value of $R_{\mathcal{N},p}(s,a)$ is either strictly negative (when a violation takes place) or zero (when the action is compliant), $Q_{\mathcal{N},p}(s,a)\le 0$ for all state-action pairs. Therefore, if we set the threshold $C_{\mathcal{N},p} = 0$, we get $CQ_{\mathcal{N},p}(s,a)=Q_{\mathcal{N},p}(s,a)$. This means that any policy $\pi$ that only follows actions from $eth(s)$ fulfills Definition 3.3. Likewise, any policy $\pi^{*}$ that only follows actions from $opt(s)$ will fulfill Definition 3.4 for an ethical-optimal policy. \qedsymbol

\subsection{Features of NGRL}
In the above subsections, we have referenced $R_{\mathcal{N},p}(s,a)$ without specifying a penalty $p$. As promised, we now address the magnitude of $p$. 
In order to address the issue of what $p$ should actually be, in practise, we will need the following lemma:

\begin{lemma}
Let $\mathcal{M}=\langle S, A, R, P\rangle$ be an MDP, with two compliance MDPs $\mathcal{M}_{\mathcal{N},p}=\langle S, A, (R, R_{\mathcal{N},p})^{T}, P\rangle$ and $\mathcal{M}_{\mathcal{N},q}=\langle S, A, (R, R_{\mathcal{N},q})^{T}, P\rangle$. Then for any policy $\pi\in \Pi_{\mathcal{M}}$: $$V^{\pi}_{\mathcal{N},p}(s) = c\cdot  V^{\pi}_{\mathcal{N},q}(s)$$ for some positive coefficient $c$, where $V^{\pi}_{\mathcal{N},p}(s)$ and $V^{\pi}_{\mathcal{N},q}(s)$ are the value functions associated with $R_{\mathcal{N},{p}}$ and $R_{\mathcal{N},q}$ respectively.
\end{lemma}
\vspace{-0.1em}
\noindent \textit{Proof sketch:} it follows directly the linearity of conditional expectation that the above property holds for the value $c=\dfrac{p}{q}\in \mathbb{R}^{+}$.
 \qedsymbol

With this lemma it is a simple matter to prove the following useful feature of NGRL:

\begin{theorem}
If a policy $\pi$ is ethical for the compliance MDP $\mathcal{M}_{\mathcal{N},p}$ for some constant $p\in \mathbb{R}^{-}$, it is ethical for the compliance MDP $\mathcal{M}_{\mathcal{N},q}$ for any $q\in \mathbb{R}^{-}$.
\end{theorem}
\noindent \textit{Proof:} Let $p\in \mathbb{R}^{-}$ be a penalty and $\pi^{*}$ be an ethical policy for $\mathcal{M_{N}}=\langle S, A, (R_{x}, R_{\mathcal{N},p})^{T}, P\rangle$. Then we know that $$V_{\mathcal{N},p}^{\pi^{*}}(s)=\max_{\pi\in \Pi_{\mathcal{M}}}V^{\pi}_{\mathcal{N},p}(s)$$ However, for any arbitrary penalty $q\in \mathbb{R}^{-}$, there is a $c\in \mathbb{R}^{+}$ such that $V_{\mathcal{N},p}^{\pi}(s)=c\cdot V_{\mathcal{N},q}^{\pi}(s)$ for any $\pi\in \Pi_{\mathcal{M}}$. Therefore
$$V_{\mathcal{N},p}^{\pi^{*}}(s)=\max_{\pi\in \Pi_{\mathcal{M}}}V^{\pi}_{\mathcal{N},p}(s)$$, 
$$c\cdot V_{\mathcal{N},q}^{\pi^{*}}(s)=\max_{\pi\in \Pi_{\mathcal{M}}}c\cdot V^{\pi}_{\mathcal{N},q}(s)$$
$$V_{\mathcal{N},q}^{\pi^{*}}(s)=\max_{\pi\in \Pi_{\mathcal{M}}}V^{\pi}_{\mathcal{N},q}(s)$$.

So if $\pi^{*}$ is an ethical policy for the compliance MDP $\mathcal{M}_{\mathcal{N},p}=\langle S, A, (R_{x}, R_{\mathcal{N},p})^{T}, P\rangle$ it is ethical for the compliance MDP $\mathcal{M}_{\mathcal{N},q}=\langle S, A, (R_{x}, R_{\mathcal{N},q})^{T}, P\rangle$ as well. 
\qedsymbol

Thus, the value of $p$ is irrelevant, so for the remainder of the paper we will only deal with the non-compliance function $R_{\mathcal{N}} := R_{\mathcal{N},-1}$. This solves the second essential challenge we identified for reward specification.

\subsection{Limitations of NGRL}
NGRL may fail to capture some desired subtleties in the form of \textit{normative conflicts} and \textit{contrary-to-duty obligations}; this is discussed below.

DDL facilitates non-monotonic reasoning, which makes it suitable for handling normative systems that may at first glance seem to be inconsistent. Take, for instance, a normative system $\mathcal{N}=\langle \mathcal{C}, \mathcal{R},>\rangle$, where $\mathcal{R}=\{\textbf{O}(b|a), \textbf{F}(b|a)\}$ and $>=\{(\textbf{O}(b|a),\textbf{F}(b|a))\}$. In DDL, we would represent the rules as $r_{1}:\ a\Rightarrow_{O}b$ and  $r_{2}:\ a\Rightarrow_{O}\neg b$,
where $r_{1}>r_{2}$. If we assume as a fact $a$, we will be able to derive the conclusion: $+\partial_{O}b$. This dictates the behaviour we would desire from the system -- the agent sees to it that $b$, or else is punished. However, $r_{2}$ (that is, $\textbf{F}(b|a)$) is still violated by this course of action, and we might want to assign the agent a (albeit lesser) punishment for this violation, incentivising the agent to avoid situations where these conflicting norms are triggered at all. In our framework, however, taking action $b$ is considered, for all intents and purposes, compliant. 
In some cases, where the conflict occurs for reasons completely out of the agent's control, our framework is the more appropriate approach. To illustrate, consider that (1) you are forbidden from surpassing the speed limit, and (2) if your child is dying and you are driving them to the hospital, you ought to speed -- then there is no reason to punish you for speeding to save your dying child, even though you have violated norm (1). 
However, in cases where the agent's own actions have resulted in a scenario where it cannot obey all applicable norms, we might want to assign punishments to any action the agent takes, albeit of different magnitudes. Differentiating between these two kinds of conflicts is a research question we hope to pursue further.

A second, more problematic scenario can be found in contrary-to-duty obligations. Contrary-to-duty obligations are obligations that come into force when another norm is violated. The classic example can be found in \cite{C1963}. Below, consider a normative system $\mathcal{N}=\langle \mathcal{C}, \mathcal{R}, >\rangle$, where $\mathcal{R}=\{\textbf{O}(b|\neg a), \textbf{O}(a|\top)\}$. In DDL these norms are: $r_{1}:\ \Rightarrow_{O} a$ and $r_{2}:\ \neg a\Rightarrow_{O} b$. 
Suppose further that the agent has 3 actions available to it: $a$, $b$, $c$. The first problem arises in that $r_2$ is triggered by the agent not taking action $a$, and when we construct $Th(s,\mathcal{N})$, this information is not included. This can be remedied; we construct a new theory which takes into account which action is taken. However, a new problem then arises;  
the obligation of $a$ along with the non-concurrence constraints over $a$ imply the prohibition of $b$ and $c$. These constraints will be strict rules, and will therefore defeat any defeasible rules with conflicting consequent. In other words, $r_2$, even if triggered, will be defeated and we will be able to derive both $+\partial_{O}\neg b$ and $+\partial_{O}\neg c$. Thus, the agent will be assigned identical punishments for performing actions $b$ and $c$, though it may be desirable to incentivize the agent to, given the choice between $b$ and $c$, choose $b$.

Based on examples like those above, we might find it appropriate to assign graded penalties to violations of a normative system, depending on how much of it is violated. 
It remains to be seen if an agent can learn how to handle cases like this without the further interference of the normative supervisor.

\section{Experimental Results}
We now review an evaluation of the methods discussed in Sect.~\ref{sect:ngql}. In Sect.~4.1 we will describe our agent's environment and the normative system it is subject to, and in 4.2 we discuss the results of training an agent using the methods above for 9000 episodes and testing over 1000.\footnote{Tests were run on a laptop with a Intel i7-7500U CPU (4 cores, 2.70  GHz) and 8GB RAM,  Ubuntu 16.04, Java 8, Python 2.7. Tests can be reproduced with the code located at \url{https://github.com/lexeree/normative-player-characters}.}  

In the below tests, we use the normative supervisor in two different ways: its original mode of operation, performing online compliance checking during test games (when it is in use, this will be designated by a \textit{yes} in the \textbf{Monitored?} column of the table in Sect.~4.2), and to facilitate NGRL.

\subsection{Benevolent Pac-Man}
\label{sect:pacman}
In \cite{NBCG2021}, the performance of the supervisor used in conjunction with a Q-learning agent with function approximation was evaluated with a simulation of the arcade game Pac-Man. Since most of the work in this paper is built on Q-learning \textit{without} function approximation, the agents we discuss cannot handle such a complex environment, so we have made a simplified version of the game for them to play. 
\begin{figure}[ht]
\label{fig:fig1}
\centering
    \includegraphics[width=0.4\columnwidth]{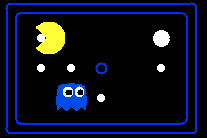}
    \caption{A miniaturized Pac-Man game where only one ghosts is present.}
\end{figure}

In this simplified game Pac-Man is located on a small $5\times 3$ grid with only a 
single ghost and power pellet located in opposite corners. Pac-Man gains 500 points if he wins the game (by eating all the food pellets), and loses 500 if he loses the game (by being eaten by the ghost, which moves randomly). There are 11 food pellets in this simplified game and Pac-Man loses 1 point per time step; thus, the maximum score (without eating the ghost) is 599 points. When Pac-Man eats the power pellet and the ghost becomes scared, Pac-Man can eat the ghost for an additional 200 points.


The normative system we use is built around the imposition of the ``duty" of ``benevolence" onto Pac-Man. This is represented by the norm 
\begin{equation}
    \textbf{O}(benevolent|\top)\in \mathcal{R}
\end{equation}
which means that Pac-Man should be benevolent at all times.
What does it mean for Pac-Man to be benevolent? We don't have an explicit definition, but we have a description of non-benevolent behaviour: 
\begin{equation}
    \textbf{C}(eat(person), \neg benevolent)\in \mathcal{C}
\end{equation}
where $eat(person)$ is a description of the act of eating an entity designated as a person. That is, we know that Pac-Man is \textit{not} being benevolent if he eats a person. Additional constitutive norms could expand our definition of benevolence. We next add a second constitutive norm, 
\begin{equation}
   \textbf{C}(eat(blueGhost), eat(person))\in \mathcal{C} 
\end{equation}
asserting that eating a blue ghost counts as eating a person. 

 This concept of eating the ghost also needs to be further specified in our normative system. We use the same formalization used in \cite{NBCG2021}, with additional constitutive norms asserting that moving into the same cell as a ghost while the ghost is scared counts as eating the ghost. For a full description of the structure of constitutive norms required to establish these concepts, see \cite{NBCG2021}. This normative system produces the same behaviour as the ``vegan" norm set used in \cite{NBCG2021} -- that is, it manifests as a prohibition from eating the blue ghost -- the difference lies in how it is formalized, using a more abstract set of concepts to govern behaviour.

\subsection{Results}
We ran the tests summarized in Table 1 on three kinds of agents: a regular Q-learning agent, and the two NGRL agents we described in Sect.~\ref{sect:ngql} (the linear scalarization agent and the TLQ-learning agent).

\begin{table}[!ht]
\caption{Results with 3 different agents, with and without normative supervision}\label{tab:1}
\centering
\scalebox{0.76}{
\begin{tabular}{|c|c|c|c|c|}
\hline
\multicolumn{1}{|c|}{\small{\textbf{Agent}}} & \small{\textbf{Monitored?}} & \small{\textbf{\% Games}}    & \small{\textbf{Avg Game}}   & \small{\textbf{Avg Ghosts}}   \\ 
\multicolumn{1}{|c|}{} & & \small{\textbf{Won}} & \small{\textbf{Score}}   & \small{\textbf{Eaten}}  \\ 
\hline
 Q-learning   & no    & 68.5\%         &  441.71 &  0.851    \\ \hline
Scalarized  &  no  &  82.0\%        &  433.74 &    0.142  \\ \hline
TLQL  &    no & 82.0\%        &  433.74 &    0.142  \\ \hline
Q-learning  & yes    &    46.6\%     & 25.22 &    0.0  \\ \hline
Scalarized  &  yes  &    86.3\%     & 448.23  &    0.001 \\ \hline
TLQL  & yes &   86.3\%     & 448.23  &   0.001  \\ \hline
\end{tabular}
}
\end{table}

The results of Table 1 show that for NGRL (without supervision), the agents failed to comply with the normative system about as often as they failed to win the game. 
It is notable that both NGRL methods, linear scalarization and TLQL, converged to the same policy, resulting in identical numbers. However, it is also worth noting that though the number of ghosts eaten was reduced significantly (in the linear scalarization and TLQL tests, the number of ghosts eaten is only one sixth of what it was for the Q-learning agent), the behaviour of eating ghosts was not even close to being eliminated. When using the normative supervisor with regular Q-learning, on the other hand, no ghosts were eaten
; in the meantime, the agent's ability to play the game was hampered greatly.

We received the best overall results when combining NGRL with normative supervision; it was in these tests that we saw the highest win rates/score, and an elimination of ghosts consumed.\footnote{An elimination with the exception of 1. We confirmed this to be the edge case described in \cite{NBCG2021}, where the agent consumes a ghost `by accident', when it collides with the ghost right as it eats the power pellet.} While the normative supervisor, in its original monitoring role, eliminates the possibility of unnecessarily violating the norm base, it does not teach the agent to learn an optimal policy \textit{within} the bounds of compliant behaviour. Consider the following example: an autonomous vehicle is travelling down a road that runs through private property, which it is forbidden from entering; if the agent is told as soon as it encounters this private property that it is forbidden from proceeding, it must turn back and find another route, while it would have been more efficient for the agent to select a compliant route from the beginning. NGRL allows the agent to incorporate information about which actions are compliant and which are not into its learning of optimal behaviour; as a result, the performance of the agent in accomplishing its non-ethical goals is not compromised to the same degree it might be when we employ only online compliance-checking.

As a final consideration, suppose we introduce a new regulative norm, a permissive norm: 
\begin{equation}
    \textbf{P}(eat(blueGhost)|\top)\in \mathcal{R}
\end{equation}
This gives Pac-Man explicit permission to eat the blue ghost, overriding any prohibitions implied by the obligation (5) and constitutive norms (6) and (7) described in Sect. 4.1. As we can see from the results in Table 2, Pac-Man will return to eating ghosts after being given explicit permission, exhibiting identical behaviour to the unmonitored Q-learning agent.

\begin{table}[!ht]
\caption{Results with Pac-Man permitted to eat the ghost.}\label{tab:2}
\centering
\scalebox{0.73}{
\begin{tabular}{|c|c|c|c|c|c|}
\hline
\multicolumn{1}{|c|}{\small{\textbf{Agent}}} & \small{\textbf{Monitored?}} & \small{\textbf{\% Games}}    & \small{\textbf{Game Score}}   & \small{\textbf{Ghosts Eaten }}   \\ 
\multicolumn{1}{|c|}{} & & \small{\textbf{Won}} & \small{\textbf{(Avg)}}   & \small{\textbf{(Avg)}}  \\ 
\hline
Scalarized  &  no  &   68.5\%   &  441.71 &  0.851   \\ \hline
TLQL  &  no   & 68.5\%      &  441.71 & 0.851    \\ \hline
\end{tabular}
}
\end{table}

\subsubsection{NGRL with Function Approximation}
Q-learning and related techniques are susceptible to inadequacy in large state spaces; this is in part because Q-values must be stored with explicit state representations in a look-up table that can become explosively large in complex environments. For this reason, function approximation is a popular way to adapt Q-learning to larger, more complex environments.

We will consider here the simple case of linear function approximation; that is, approximating the Q-function as a linear combination of features. For this technique, we want to find a weight vector $\vec{\theta}$ for a 
vector of functions $f_{i}(s,a):S\times A\to \mathbb{R}$ extracting features from the environment. What the approximated Q-function looks like, then, is a linear combination of features: $$Q_{\theta}(s,a)=\theta_{1}f_{1}(s,a)+...+\theta_{n}f_{n}(s,a)$$ with each $f_{i}(s,a)$ extracting some feature of the environment (e.g., in state $s$ the agent is $f_{i}(s,a)$ steps away from object $x$) and each $\theta_{i}$ being a learned weight on this feature. In this section, we go over the results of approximating $Q_{x}$ and $Q_{\mathcal{N}}$ (separately) with such linear functions, after which they are once again scalarized with the method used in 3.1. 

In our case, using function approximation allows us to model the agent's behaviour in a much larger environment, shown in Fig. 2.
\begin{figure}[ht]
\centering
    \includegraphics[width=\columnwidth]{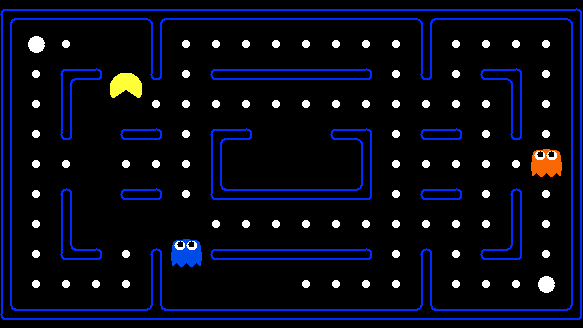}
    \caption{Pac-Man in a larger maze with two ghost opponents.}
\end{figure}
In this environment, there are two ghosts, and it is possible for Pac-Man to be trapped between them, resulting in his death (if the ghosts are not scared) or the death of one or more ghost (if the ghosts are scared).

Our first tests were within the same normative system employed in 4.2, prior to the addition of the permissive norm (8). The only difference is that we added a constitutive norm for the orange ghost:
\begin{equation}
    \textbf{C}(eat(orangeGhost), eat(person))\in \mathcal{C}
\end{equation}
The results in Table 3 were obtained from batches of 1000 tests, performed after Pac-Man had been trained over 300 games. Here, training entails adjusting the weight vector $\vec{\theta}$ to create a better approximation of the Q-function. 
Note that the below results do not include function approximation with TLQ-learning; this technique strictly prioritizes the ethical objective, and so long as we maintain the threshold of 0 for $Q_{\mathcal{N}}$, Pac-Man stayed still in his initial position, where there was no chance of him violating his normative system.

\begin{table}[!ht]
\caption{Results for tests on NGRL with function approximation.}\label{tab:3}
\centering
\scalebox{0.7}{
\begin{tabular}{|c|c|c|c|c|c|c|}
\hline
\multicolumn{1}{|c|}{\small{\textbf{Agent}}} & \small{\textbf{Monitored?}} &  \small{\textbf{\% Games}}    & \small{\textbf{Game Score}}   & \small{\textbf{Avg ghosts eaten }}   \\ 
\multicolumn{1}{|c|}{} &  &  \small{\textbf{Won}} & \small{\textbf{(Avg)}}   & \small{\textbf{(Blue / Orange)}}  \\ 
\hline
 Q-learning & no &   90.8\%   &  1545.91   &   0.823 / 0.846 \\ \hline
Scalarized & no &     91.9\%       &   1228.40   &  0.019   / 0.015 \\ \hline
 Q-learning & yes &   91.1\%    &    1216.41 &   0.002 / 0.041 \\ \hline
Scalarized & yes &     90.4\%       &  1204.47    &    0.0 / 0.021 \\ \hline
\end{tabular}
}
\end{table}

From these results we can see that the NGRL approach is more effective than the monitored Q-learning agent; this is presumably because of the scenarios where Pac-Man becomes trapped between two scared ghosts, and is forced to eat one of them. 
Using NGRL, Pac-Man will learn to avoid some of these situations, while the monitored Q-learning agent must be reactive in its behaviour. However, we can also see that once again the combined use of NGRL and supervision is the most effective, suggesting that these two approaches should be used complementarily.

However, problems arise when we adopt the permissive norm (8) introduced in 4.2. The agent does not take the permission into account, and proceeds to learn as though we had not added this norm
-- this is because of the feature extractor used for the Q-function approximation, which was the same as the one used in \cite{NBCG2021}. 
\begin{table*}[!ht]
\caption{Results with permission norm.}\label{tab:4}
\centering
\begin{tabular}{|c|c|c|c|c|c|c|c|}
\hline
\multicolumn{1}{|c|}{\small{\textbf{Agent}}} & \small{\textbf{Monitored?}} & \small{\textbf{Feature}} & \small{\textbf{\% Games}}    & \small{\textbf{Game Score}}   & \small{\textbf{Avg ghosts eaten }}   \\ 
\multicolumn{1}{|c|}{} &  & \small{\textbf{Extractor}} & \small{\textbf{Won}} & \small{\textbf{(Avg)}}   & \small{\textbf{(Blue / Orange)}}  \\ 
\hline
linear scalarization & no &  basic   &      91.7\%       & 1226.41     &    0.018 / 0.017 \\ \hline
 Q-learning & yes & basic &          89.1\%  &  1351.20  &  0.831 / 0.0 \\ \hline
 linear scalarization & no &  blue  &            94.4\% &  1423.87    &    0.813 / 0.020 \\ \hline
\end{tabular}
\end{table*}
The original extractor, which we call `basic' in Table 4, does not differentiate between blue and orange ghosts, and instead uses as a feature the number of ghosts that are within one step of Pac-Man. As a result, Pac-Man cannot learn to stop eating only the blue ghosts -- there is no way to distinguish states where a blue ghost is near Pac-Man from ones where an orange ghost is. However, we can split this into two features, one giving 1 if a blue ghost is nearby and 0 otherwise, and the other doing the same for the orange ghost. We called this the 'blue' extractor, and it led to the final results in Table 4.

Though Pac-Man certainly did display the behaviour of avoiding the consumption of orange ghosts, he still did eat 20 orange ghosts over 1000 games. Clearly, though NGRL can better avoid violations of some normative systems than a monitored Q-learning agent, it at the same time is prone to unnecessary violations -- suggesting, again, that these two methods for constraining behaviour are best used together. That is, the agent can learn ethical behaviour, while still having a supervisor correct for any avoidable violations that are ignored by the agent's policy, and recording any violations that do occur for the sake of transparency. This is where we get our best results. 

It may even be possible to improve the agent's ability to avoid situations resulting in unavoidable violations as occurred with respect to $\mathcal{N}$, provided that the correct features are selected. Unfortunately, doing so requires manual work involving comparing domain knowledge of the agent's environment and the normative systems, which could be an unwieldy or downright unfeasible task in more complex cases. Summarily, scaling NGRL through linear function approximation would require significant further investigation into the problem of feature engineering.

\section{Conclusion}
Teaching autonomous agents ethically or legally compliant behaviour through reinforcement learning entails assigning penalties to behaviour that violates whatever normative system the agent is subject to. To do so, we must determine (1) when to assign a penalty, and (2) how steep that penalty should be. We have addressed both of these challenges with norm-guided reinforcement learning (NGRL), a framework for learning that utilizes a normative supervisor that assesses the agent's actions with respect to a normative system. In doing so, we relegate (1) to the assessments of the normative supervisor, and sidestep (2) with techniques that produce ethical policies, showing that they will do so regardless of the magnitude of punishment given.
Our approach provides us with a 
framework for identifying non-compliant actions with respect to a given normative system, and assigning punishments to a MORL agent simultaneously learning to achieve ethical and non-ethical objectives. 

NGRL offers more versatility with respect to the complexity of the norms to be adhered to, than directly assigning rewards to specific events or with respect to simple constraints. It may be the case that there is no obvious or coherent way to summarize an entire normative system by selecting specific events and assigning punishments to them. By using NGRL, 
we expand what kinds of normatively compliant behaviour we can learn, and are allowed to specify them in a more natural way.

Our experimental results showed that NGRL was effective in producing an agent that learned to avoid most violations -- even in a stochastic environment -- while still pursuing its non-ethical goal. However, these results also revealed that we achieve optimal results when we use NGRL in conjunction with the normative supervisor as originally intended, as an online compliance-checker. NGRL allows us to circumvent the weaknesses of the normative supervision approach -- namely, its inability to preemptively avoid violations -- while normative supervision allows us to maintain a better guarantee of compliance.

As discussed in Sect.~3.3.1, NGRL can be further developed in its handling of normative conflict and contrary-to-duty obligations. Moreover, as this approach applies only to MORL variants of Q-learning, it will fall prey to the same scaling issues. Adapting NGRL to be used with Q-learning with function approximation, for example, would broaden the domains to which NGRL can applied.

\section*{\uppercase{Acknowledgements}}
This work was supported by the DC-RES run by the TU Wien’s Faculty of Informatics and the FH-Technikum Wien.

\bibliographystyle{apalike}
{\small
\bibliography{example}}

\begin{thebibliography}{}

\bibitem[Abel et~al., 2016]{AML2016}
Abel, D., MacGlashan, J., and Littman, M.~L. (2016).
\newblock Reinforcement learning as a framework for ethical decision making.
\newblock In {\em AAAI Workshop: AI, Ethics, and Society}, volume~16.

\bibitem[Alshiekh et~al., 2018]{ABEKNT2018}
Alshiekh, M., Bloem, R., Ehlers, R., K{\"o}nighofer, B., Niekum, S., and Topcu,
  U. (2018).
\newblock Safe reinforcement learning via shielding.
\newblock In {\em Proceedings of the AAAI Conference on Artificial
  Intelligence}, volume~32.

\bibitem[Balakrishnan et~al., 2019]{BBMR2019}
Balakrishnan, A., Bouneffouf, D., Mattei, N., and Rossi, F. (2019).
\newblock Incorporating behavioral constraints in online {AI} systems.
\newblock In {\em Proceedings of the AAAI Conference on Artificial
  Intelligence}, volume~33, pages 3--11.

\bibitem[Barrett and Narayanan, 2008]{BN2008}
Barrett, L. and Narayanan, S. (2008).
\newblock Learning all optimal policies with multiple criteria.
\newblock In {\em Proceedings of the 25th international conference on Machine
  learning}, pages 41--47.

\bibitem[Boella and van~der Torre, 2003]{BT2003}
Boella, G. and van~der Torre, L. (2003).
\newblock Permissions and obligations in hierarchical normative systems.
\newblock In {\em ICAIL}.

\bibitem[Boella and van~der Torre, 2004]{BT2004}
Boella, G. and van~der Torre, L. (2004).
\newblock Regulative and constitutive norms in normative multiagent systems.
\newblock In {\em Proc. of KR 2004: the 9th International Conference on
  Principles of Knowledge Representation and Reasoning}, pages 255--266. {AAAI}
  Press.

\bibitem[Broersen et~al., 2013]{BCEGG2013}
Broersen, J.~M., Cranefield, S., Elrakaiby, Y., Gabbay, D.~M., Grossi, D.,
  Lorini, E., Parent, X., van~der Torre, L. W.~N., Tummolini, L., Turrini, P.,
  and Schwarzentruber, F. (2013).
\newblock Normative reasoning and consequence.
\newblock In {\em Normative Multi-Agent Systems}, volume~4 of {\em Dagstuhl
  Follow-Ups}, pages 33--70. Schloss Dagstuhl - Leibniz-Zentrum f{\"{u}}r
  Informatik.

\bibitem[Chisholm, 1963]{C1963}
Chisholm, R.~M. (1963).
\newblock Contrary-to-duty imperatives and deontic logic.
\newblock {\em Analysis}, 24(2):33--36.

\bibitem[G{\'a}bor et~al., 1998]{GKS1998}
G{\'a}bor, Z., Kalm{\'a}r, Z., and Szepesv{\'a}ri, C. (1998).
\newblock Multi-criteria reinforcement learning.
\newblock In {\em Proceedings of the Fifteenth International Conference on
  Machine Learning.}, volume~98, pages 197--205.

\bibitem[Garc{\i}a and Fern{\'a}ndez, 2015]{GF2015}
Garc{\i}a, J. and Fern{\'a}ndez, F. (2015).
\newblock A comprehensive survey on safe reinforcement learning.
\newblock {\em Journal of Machine Learning Research}, 16(1):1437--1480.

\bibitem[Governatori, 2018]{G2018}
Governatori, G. (2018).
\newblock Practical normative reasoning with defeasible deontic logic.
\newblock In {\em Reasoning Web International Summer School}, pages 1--25.
  Springer.

\bibitem[Governatori and Hashmi, 2015]{GH2015}
Governatori, G. and Hashmi, M. (2015).
\newblock No time for compliance.
\newblock In {\em IEEE 19th International Enterprise Distributed Object
  Computing Conference}, pages 9--18. IEEE.

\bibitem[Governatori et~al., 2013]{GORS2013}
Governatori, G., Olivieri, F., Rotolo, A., and Scannapieco, S. (2013).
\newblock Computing strong and weak permissions in defeasible logic.
\newblock {\em Journal of Philosophical Logic}, 42(6):799--829.

\bibitem[Governatori and Rotolo, 2008]{jaamas:bio}
Governatori, G. and Rotolo, A. (2008).
\newblock {BIO} logical agents: Norms, beliefs, intentions in defeasible logic.
\newblock {\em Journal of Autonomous Agents and Multi Agent Systems},
  17(1):36--69.

\bibitem[Hasanbeig et~al., 2019]{HAK2018}
Hasanbeig, M., Abate, A., and Kroening, D. (2019).
\newblock Logically-constrained neural fitted q-iteration.
\newblock In {\em Proceedings of the 18th International Conference on
  Autonomous Agents and MultiAgent Systems}, AAMAS '19, page 2012–2014.

\bibitem[Hasanbeig et~al., 2020]{HAK2020}
Hasanbeig, M., Abate, A., and Kroening, D. (2020).
\newblock Cautious reinforcement learning with logical constraints.
\newblock In {\em Proceedings of the 19th International Conference on
  Autonomous Agents and Multiagent Systems, {AAMAS} '20, Auckland, New Zealand,
  May 9-13, 2020}, pages 483--491.

\bibitem[{Hasanbeig} et~al., 2019]{HKAKPL2019}
{Hasanbeig}, M., {Kantaros}, Y., {Abate}, A., {Kroening}, D., {Pappas}, G.~J.,
  and {Lee}, I. (2019).
\newblock Reinforcement learning for temporal logic control synthesis with
  probabilistic satisfaction guarantees.
\newblock In {\em Proc. of {CDC} 2019: the 58th {IEEE} Conference on Decision
  and Control}.

\bibitem[Jansen et~al., 2020]{JKJSB2020}
Jansen, N., K{\"o}nighofer, B., Junges, S., Serban, A., and Bloem, R. (2020).
\newblock {Safe Reinforcement Learning Using Probabilistic Shields (Invited
  Paper)}.
\newblock In {\em 31st International Conference on Concurrency Theory (CONCUR
  2020)}, volume 171 of {\em Leibniz International Proceedings in Informatics
  (LIPIcs)}, pages 3:1--3:16.

\bibitem[Jones and Sergot, 1996]{JS1996}
Jones, A. J.~I. and Sergot, M. (1996).
\newblock {A Formal Characterisation of Institutionalised Power}.
\newblock {\em Logic Journal of the IGPL}, 4(3):427--443.

\bibitem[Kasenberg and Scheutz, 2018]{KS2018}
Kasenberg, D. and Scheutz, M. (2018).
\newblock Norm conflict resolution in stochastic domains.
\newblock In {\em Proceedings of the AAAI Conference on Artificial
  Intelligence}, volume~32.

\bibitem[Lam and Governatori, 2009]{ruleml09:spindle}
Lam, H.-P. and Governatori, G. (2009).
\newblock The making of {SPIN}dle.
\newblock In {\em Proc. of RuleML 2009: International Symposium on Rule
  Interchange and Applications}, volume 5858 of {\em LNCS}, Heidelberg.
  Springer.

\bibitem[Neufeld et~al., 2021]{NBCG2021}
Neufeld, E., Bartocci, E., Ciabattoni, A., and Governatori, G. (2021).
\newblock A normative supervisor for reinforcement learning agents.
\newblock In {\em Proceedings of {CADE} 28 - 28th International Conference on
  Automated Deductions}, pages 565--576.

\bibitem[Noothigattu et~al., 2019]{N2019}
Noothigattu, R., Bouneffouf, D., Mattei, N., Chandra, R., Madan, P., Varshney,
  K.~R., Campbell, M., Singh, M., and Rossi, F. (2019).
\newblock Teaching {AI} agents ethical values using reinforcement learning and
  policy orchestration.
\newblock In {\em Proc of IJCAI: 28th International Joint Conference on
  Artificial Intelligence}.

\bibitem[Palmirani et~al., 2011]{PGRTBP2011}
Palmirani, M., Governatori, G., Rotolo, A., Tabet, S., Boley, H., and Paschke,
  A. (2011).
\newblock Legalruleml: Xml-based rules and norms.
\newblock In {\em International Workshop on Rules and Rule Markup Languages for
  the Semantic Web}, pages 298--312. Springer.

\bibitem[Rodriguez-Soto et~al., 2021]{RLR2021}
Rodriguez-Soto, M., Lopez-Sanchez, M., and Rodriguez~Aguilar, J.~A. (2021).
\newblock Multi-objective reinforcement learning for designing ethical
  environments.
\newblock In {\em Proceedings of the Thirtieth International Joint Conference
  on Artificial Intelligence, {IJCAI-21}}, pages 545--551. International Joint
  Conferences on Artificial Intelligence Organization.

\bibitem[Roijers et~al., 2013]{RVWD2013}
Roijers, D.~M., Vamplew, P., Whiteson, S., and Dazeley, R. (2013).
\newblock A survey of multi-objective sequential decision-making.
\newblock {\em Journal of Artificial Intelligence Research}, 48:67--113.

\bibitem[Vamplew et~al., 2011]{VDBID2011}
Vamplew, P., Dazeley, R., Berry, A., Issabekov, R., and Dekker, E. (2011).
\newblock Empirical evaluation methods for multiobjective reinforcement
  learning algorithms.
\newblock {\em Machine learning}, 84(1):51--80.

\bibitem[Vamplew et~al., 2018]{VDFFM2018}
Vamplew, P., Dazeley, R., Foale, C., Firmin, S., and Mummery, J. (2018).
\newblock Human-aligned artificial intelligence is a multiobjective problem.
\newblock {\em Ethics and Information Technology}, 20(1):27--40.

\bibitem[Wu and Lin, 2018]{WL2018}
Wu, Y.-H. and Lin, S.-D. (2018).
\newblock A low-cost ethics shaping approach for designing reinforcement
  learning agents.
\newblock In {\em Proceedings of the AAAI Conference on Artificial
  Intelligence}, volume~32.

\end{thebibliography}

\end{document}